\documentclass[runningheads]{llncs}

\usepackage{eccv}
\usepackage{diagbox}

\usepackage{eccvabbrv}

\usepackage{graphicx}
\usepackage{booktabs}
\usepackage{color}

\usepackage[accsupp]{axessibility}  

\usepackage[pagebackref,breaklinks,colorlinks,citecolor=eccvblue]{hyperref}

\usepackage{orcidlink}
\usepackage{algorithmic}
\usepackage{algorithm}
\usepackage{amsmath}
\usepackage{amssymb}
\usepackage{amsfonts,amssymb}
\usepackage{amsmath}
\usepackage{color}
\usepackage{bbding}
\usepackage{multirow}
\usepackage{makecell}
\usepackage{float}
\usepackage{multicol}
\usepackage{tablefootnote}

\usepackage{ulem}

\renewcommand{\sout}[1]{}

\begin{document}

\title{Unsupervised Moving Object Segmentation \\with Atmospheric Turbulence} 


\author{Dehao Qin\inst{1} \and
Ripon Kumar Saha\inst{2} \and
Woojeh Chung\inst{2} \and
Suren Jayasuriya\inst{2} \and
Jinwei Ye\inst{3} \and
Nianyi Li\inst{1}}

\authorrunning{Qin et al.}

\institute{$^{1}$Clemson University, 
$^{2}$Arizona State University,
$^{3}$George Mason University \\
\url{https://turb-research.github.io/segment\_with\_turb}
}

\maketitle

\begin{abstract}
Moving object segmentation in the presence of atmospheric turbulence is highly challenging due to turbulence-induced irregular and time-varying distortions. In this paper, we present an unsupervised approach for segmenting moving objects in videos downgraded by atmospheric turbulence. Our key approach is a detect-then-grow scheme: we first identify a small set of moving object pixels with high confidence, then gradually grow a foreground mask from those seeds to segment all moving objects. This method leverages rigid geometric consistency among video frames to disentangle different types of motions, and then uses the Sampson distance to initialize the seedling pixels. After growing per-frame foreground masks, we use spatial grouping loss and temporal consistency loss to further refine the masks in order to ensure their spatio-temporal consistency. Our method is unsupervised and does not require training on labeled data. For validation, we collect and release the first real-captured long-range turbulent video dataset with ground truth masks for moving objects. Results show that our method achieves good accuracy in segmenting moving objects and is robust for long-range videos with various turbulence strengths. 

\end{abstract}

\section{Introduction}


Moving object segmentation is critical for motion understanding with an important role in numerous vision applications such as security surveillance~\cite{LING201432,Osorio2015Surveillance}, remote sensing~\cite{9157665,Rai2021Moving}, and environmental monitoring~\cite{rs12111772,Chen2021Moving}. Although tremendous success has been achieved in motion segmentation and analysis~\cite{cho2023treating,LIU2022103700}, the problem becomes highly challenging for long-range videos captured with ultra-telephoto lenses (e.g., focal length $> 800$mm). These videos often suffer from perturbations caused by atmospheric turbulence which can geometrically shift or warp pixels in images. When mixed with rigid motions in a dynamic scene, they break down the underlying assumption of most motion analysis algorithms: the intensity structures of local regions are constant under motion. Further, when averaged over time, the turbulent perturbation also yields blurriness in images, which blurs out moving object edges and makes it challenging to maintain the spatio-temporal consistency of segmentation masks.

\begin{figure}[t]
  \centering
   \includegraphics[width=0.8\linewidth]{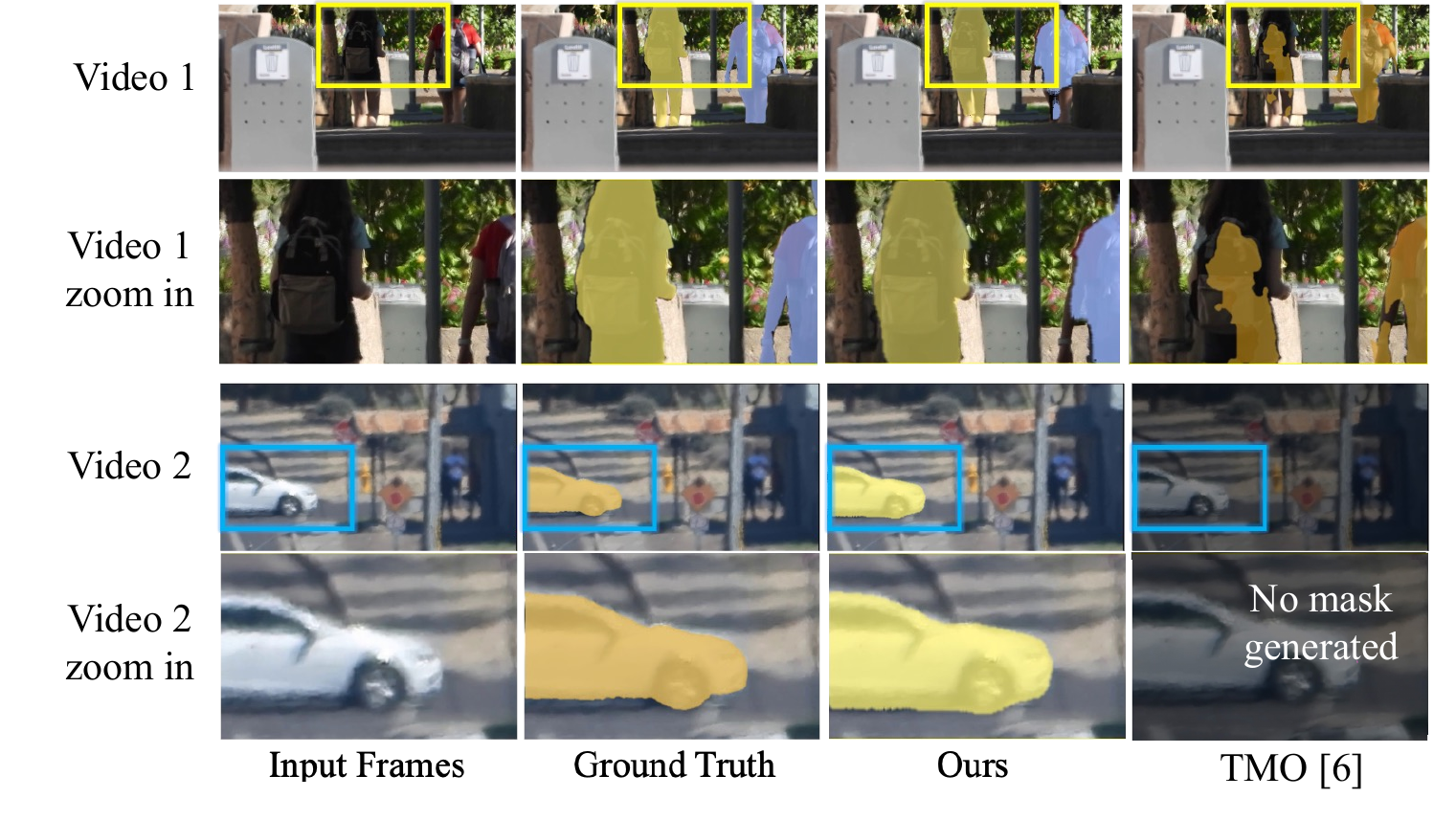}
   \caption{
   Our method robustly segments moving objects under various turbulence strengths, while state-of-the-art methods may fail under strong turbulence (2nd video). }
   \label{fig:teaser}
\vspace{-10pt}   
\end{figure}

Most motion segmentation algorithms solely consider static backgrounds and assume rigid body movement. Under these premises, learning-based approaches~\cite{cho2023treating,LIU2022103700,Wang2021A}, either supervised or unsupervised, have achieved remarkable success in motion segmentation. However, these algorithms' performance significantly downgrades when applied to videos with turbulence effects (see failure examples in Fig.~\ref{Qualitative comparison}). Supervised methods~\cite{9551804,Wang2021A,kirillov2023segment}, even trained on extensive labeled data, cannot generalize well on turbulent videos. Unsupervised methods~\cite{cho2023treating,LIU2022103700}, on the other hand, typically rely on optical flow, which becomes inaccurate when rigid motion is perturbed by turbulence. Furthermore, imaging at long distance makes videos highly susceptible to camera shake and motion due to the limited field of view when zooming, further complicating the segmentation task.

\begin{figure*}[t!]
\centering
\includegraphics[width=1\textwidth]{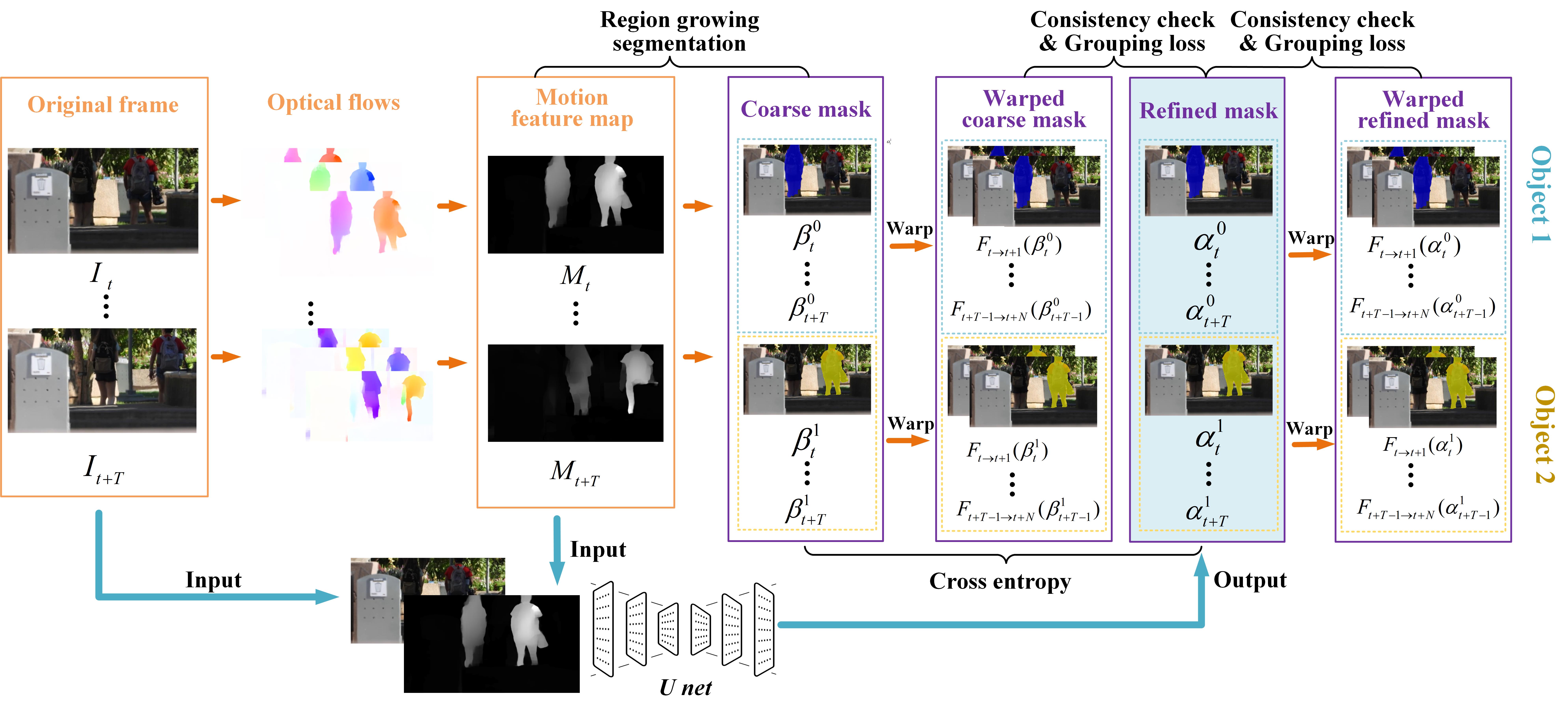} 
\caption{Overall pipeline of our unsupervised motion segmentation method. We first generate motion feature maps by applying a geometry-based consistency check on optical flows. We then adopt a region-growing scheme to generate coarse segmentation masks. Finally, we refine the masks using cross entropy-based consistency losses to enforce their spatio-temporal consistency.}
\label{pipeline}
\vspace{-10pt}
\end{figure*}

In this paper, we present an unsupervised approach for segmenting moving objects in long-range videos affected by air turbulence. The unsupervised nature of our approach is highly desirable, since real-captured turbulent video datasets are scarce and difficult to acquire. Our method directly takes in a turbulent video and outputs per-frame masks that segment all moving objects without the need for data supervision. The overall pipeline of our approach is illustrated in Fig.~\ref{pipeline}. Our method starts with calculating bidirectional optical flow. To disentangle actual object motion from turbulent motion, we use a novel epipolar geometry-based consistency check to generate motion feature maps that only preserve object motions. We then adopt a region-growing scheme that generates per-object motion segmentation masks from a small set of seed pixels. Finally, we develop a U-Net \cite{ronneberger2015unet} trained by our proposed bidirectional consistency losses and a pixel grouping function to improve the spatio-temporal consistency of estimated motion segmentation masks. 

For evaluation, we collect and release a turbulent video dataset captured with an ultra-telephoto lens called Dynamic Object Segmentation in Turbulence (DOST), and manually annotate ground-truth masks for moving objects, which, to our knowledge, is the \textit{first} dataset in this application domain. We benchmark our approach, as well as other state-of-the-art segmentation algorithms, on this real dataset. Our method is able to handle multiple objects in a dynamic scene and is robust to videos captured with various turbulence strengths (see Fig.~\ref{fig:teaser}).
More details about DOST datasest can be found in \url{https://turb-research.github.io/DOST/}.
Our key contributions include: 
\begin{itemize}
\item  A rigid geometry-based and consistency enhanced framework for motion disentanglement in long-range videos. 
\item Region-growing scheme for generating robust spatio-temporal consistent masks with tight object boundaries. 
\item A refinement pipeline with novel training losses, which improve the spatio-temporal consistency for segmenting dynamic objects in videos.
\item The \textit{first} real-captured long-range turbulent video dataset with ground-truth motion segmentation masks. 
\end{itemize}

\label{sec:intro}

\section{Related Work}

\noindent\textbf{Unsupervised Motion Segmentation.} Early approaches tackle this problem using handcrafted features such as objectness~\cite{6618931}, saliency~\cite{7298961}, and motion trajectory~\cite{6126418}.
Recent learning-based unsupervised methods largely rely on either optical flow or feature alignment for locating moving objects. Common optical flow models are RAFT~\cite{teed2020raft}, PWC-Net~\cite{sun2018pwc}, and FLow-Net~\cite{dosovitskiy2015flownet, ilg2017flownet}. MP-Net~\cite{8099547} uses optical flow as the only cue for motion segmentation. Many other methods combine optical flow with other information, such as appearance features~\cite{yang2021learning, pei2022hierarchical,lee2023unsupervised}, long-term temporal information~\cite{zhang2021deep,ren2021reciprocal}, and boundary similarity~\cite{liu2021f2net,zhou2021flow}, to guide motion segmentation. Our method also uses optical flow but segregates different types of motion using a geometry-based consistency check to overcome turbulence-induced errors.

Another class of methods~\cite{lu2020video,yang2019anchor,garg2021mask} rely more on appearance features and reduce their dependency on optical flow, so to achieve robust performance regardless of motion. The recent TMO~\cite{cho2023treating} achieves state-of-the-art performance on object segmentation by prioritizing feature alignment and fusion. However, all these methods face challenges when the input video is recorded under air turbulence. Further, the literature on supervised motion segmentation (e.g.,~\cite{li2019motion,mahadevan2020making,yin2021agunet}) is not effective due to the lack of large, labelled turbulent video datasets.

\noindent \textbf{Geometric Constraints for Motion Analysis.}
Geometric constraints are extensively studied for understanding spatial information, such as depth estimation, pose estimation, camera calibration~\cite{Zou_2018,8578310,8953778}, and 3D reconstruction~\cite{shin20193d, zhou2023sparsefusion}. They can also be useful in understanding motion. Valgaerts et al.~\cite{10.1007/978-3-540-69321-5_32} propose a variational model to estimate the optical flow, along with the fundamental matrix. Wedel et al.~\cite{5459375} use the fundamental matrix as a weak constraint to guide optical flow estimation. Yamaguchi et al.~\cite{6619087} convert the problem of optical flow estimation into a 1D search problem by using precomputed fundamental matrices with small motion assumptions. Wulff et al.~\cite{8100214} use semantic information to separate dynamic objects from static backgrounds and apply strong geometric constraints to the static backgrounds. Zhong et al.~\cite{zhong2019epiflow} integrate global geometric constraints into network learning for unsupervised optical flow estimation. More recently, Ye et al.~\cite{9880015} use the Sampson error, which measures the consistency of epipolar geometry, to model a loss function for layer decomposition in videos.
%
%

Inspired by these works, our method adopts a geometry-based consistency check to separate rigid motion from other types of motion (i.e., turbulent motion and camera motion). Similar to~\cite{9880015}, we use the Sampson distance to measure the geometric consistency between neighboring video frames. 
\vspace{2pt}

\noindent\textbf{Turbulent Image/Video Restoration and Segmentation.} Analyzing images or videos affected by air turbulence has been a challenging problem in computer vision due to distortions and blur caused by turbulence. Most techniques have focused on turbulent image and video restoration. Early physics-based approaches~\cite{gutierrez2006simulation,potvin2007parametric,first} investigate the physical modeling of turbulence (e.g., the Kolmogorov model~\cite{first}), and then invert the model to restore clear images. 
A popular class of methods~\cite{fried1978probability,Aubailly:09} uses ``lucky patches'' to reduce turbulent artifacts, although they typically assume static scenes. Motion cues, such as optical flow, are also used for turbulent image restoration \cite{6976872,4587525,of2}. Notably, Mao et al.~\cite{9216531} adopt an optical-flow guided lucky patch technique to restore images of dynamic scenes. 
%
%
More recently, neural networks have been used to restore turbulent images or videos. Mao et al.~\cite{Turbnet2022} introduce a physics-inspired transformer model for restoration, and Zhang et al.~\cite{Zhang_2024} improve this thread of work to demonstrate state-of-the video restoration. Li et al. \cite{9710477} propose an unsupervised network to mitigate the turbulence effect using deformable grids. Jiang et al. \cite{jiang2023nert} extend this deformable grid model to handle more realistic turbulence effects. 

In contrast, object segmentation with atmospheric turbulence is relatively understudied. Cui and Zhang~\cite{cui2019accurate} propose a supervised network for semantic segmentation with turbulence. They generate a physically-realistic simulated training dataset, but the method cannot handle scene motion and suffers from real-world domain generalization. Saha et al.~\cite{saha} use simple optical flow-based segmentation in their turbulent video restoration pipeline.
Unlike these existing methods, our approach aims to segment moving objects without restoring or enhancing the turbulent video. In this way, our method is able to generate segmentation masks that are consistent with the actual turbulent video. 

\section{Unsupervised Moving Object Segmentation}
\vspace{2pt}

\noindent\textbf{Algorithm Overview.} Given an input turbulent video: $\{I_{t}|t = 1, 2, \ldots, T\}$ (where $T$ is the total number of frames, and $I_{t}$ represents a frame in the video), we first calculate its bidirectional optical flow: $\mathcal{O}_t = \{ F_{t \rightarrow t\pm i}|i = 1,\ldots, B \}$ (where $B$ is the maximum number of frames used for calculation; $F_{t \rightarrow t+i}$ is the forward flow, and $F_{t \rightarrow t-i}$ is the backward flow). We then perform an epipolar geometry-based consistency check to disentangle rigid object motion from turbulence-induced motions and camera motions (Section~\ref{geo_consis}). We output per-frame motion feature maps: $\{M_{t}|t = 1, 2, \ldots, T\}$ to characterize candidate motion regions. Next, we leverage a detect-then-grow strategy, named ``region growing'', to generate motion segmentation masks: $\{{\beta}^{m}_{t}|t = 1, 2, \ldots, T\}$ for every moving object $m$, from a small set of seedling pixels selected from $\left\{M_{t}\right\}_{t=1}^T$  (Section~\ref{sec_region_grow}). Finally, we further refine the masks by using a U-Net trained with our proposed bidirectional spatial-temporal consistency losses and pixel grouping loss (Section~\ref{sec_refine}). The final output is a set of per-frame, per-object binary masks: $\{{\alpha}^{m}_{t}|t = 1, 2, \ldots, T\}$ segmenting each moving object in a dynamic scene.

\subsection{Epipolar Geometry-based Motion Disentanglement}\label{geo_consis}
\begin{figure*}[tb]
  \centering
    \includegraphics[width=1\linewidth]{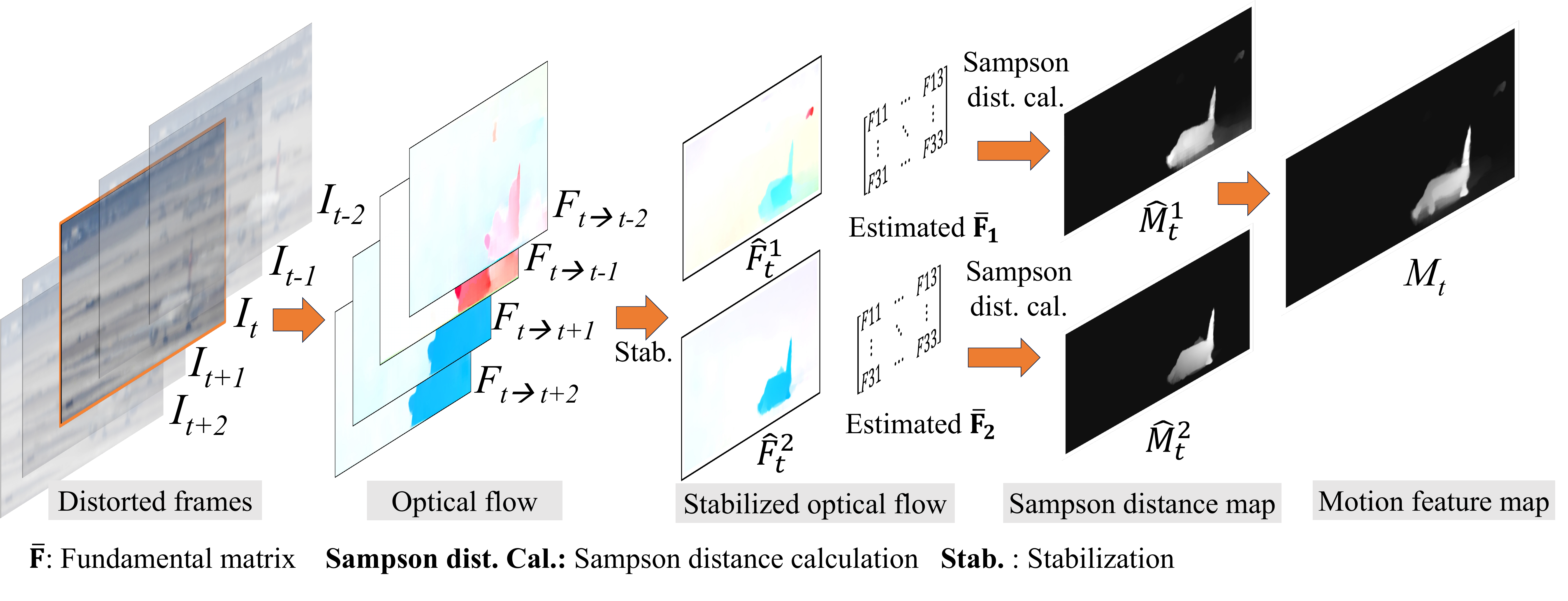}
    \caption{Pipeline of epipolar geometry-based motion disentanglement. Since the raw optical flows are downgraded by turbulence, we apply a geometry-based consistency check to generate motion feature maps that only preserve object motion.}
    \label{fig:epipolar}
\vspace{-10pt}    
\end{figure*}

We first tackle the problem of motion disentanglement, which is a major challenge posed by turbulence perturbation in rigid motion analysis. Our key idea is to check on the rigid geometric consistency among video frames: pixels on moving objects do not obey the geometric consistency constraint posed by the image formation model. Specifcially, we use the Sampson distance, which measures geometric consistency with a given epipolar geometry, to improve the spatial-temporal consistency among video frames. We first average the optical flow between adjacent frames to stabilize the direct estimations $\{\mathcal{O}_t\}$, since they are susceptible to turbulence perturbation. We then calculate the Sampson distance using fundamental matrices estimated from the averaged optical flow. Next, we merge the Sampson distance maps as the motion feature maps $\{M_{t}|t = 1, 2, \ldots, T\}$. We use the motion feature map values as indicators of how likely a pixel has rigid motion (the higher the value, the higher the likelihood). Our epipolar geometry-based motion disentanglement pipeline is shown in Fig.~\ref{fig:epipolar}.

\noindent\textbf{Optical Flow Stabilization.} Since atmospheric turbulence causes erratic pixel shifts in video frames~\cite{9216531}, estimating optical flow is prone to error. We address this by first stabilizing the optical flow estimations: we assume consistent object motion during a short period of time, and then average the optical flow within a small time step to reduce the error caused by turbulence perturbation without losing features of the actual rigid motion.
Specifically, given a sequence of bi-directional optical flow $\mathcal{O}_t=\{F_{t\rightarrow t\pm i}\}_{i=0}^B$, we calculate a sequence of per-frame stabilized flow $\hat{\mathcal{O}_t} = \{\hat{F}_t^j | j = 1, 2,..., A\}$ (where $A$ is the total number of stabilized flows for each frame) by averaging the original sequence within a short interval:
\begin{equation}
  \hat{F}_{t}^j = \frac{1}{|\mathcal{K}^j|} \sum_{i \in \mathcal{K}^j} \frac{ F_{t \rightarrow t+i}}{i},
  \label{averaged optical flow}
\end{equation}
where $\mathcal{K}^j$ is the temporal interval used for calculating $\hat{F}_t^j$, namely the subset of $\{x \mid x \in \mathbb{Z}, -B \leq x \leq B\}$

\vspace{2pt}

\noindent\textbf{Geometric Consistency Check.}
Our fundamental assumption is that pixels on moving objects have larger geometric consistency errors compared with static background, when mapping a frame to its neighboring time frame using foundamental matrix.
The Sampson distance~\cite{epi_book} measures rigid geometric consistency by calculating the distance between a frame pair in video, constrained by epipolar geometry. Since moving objects do not obey the epipolar geometry that assumes a static scene, their correspondences will have a large Sampson distance. Although the turbulent perturbation also breaks down the epipolar geometry, the errors that they introduce are much smaller and more random, and thus easily eliminated through averaging. 


Given a stabilized optical flow $\hat{F}_t^j$, we calculate its Sampson distance map $\hat{M}_t^j$ as: 
\begin{equation}
\hat{M}_t^j(\mathbf{p}_1, \mathbf{p}_2) = \frac{(\mathbf{p}_2^\mathsf{T} \mathbf{\bar{F}} \mathbf{p}_1)^2}
{(\mathbf{\bar{F}} \mathbf{p}_1)_1^2 + (\mathbf{\bar{F}} \mathbf{p}_1)_2^2 + (\mathbf{\bar{F}}^\mathsf{T} \mathbf{p}_2)_1^2 + (\mathbf{\bar{F}}^\mathsf{T} \mathbf{p}_2)_2^2},
\label{eq:sampson}
\end{equation}
where $\mathbf{p}_1$ and $\mathbf{p}_2$ are the homogeneous coordinates of a pair of corresponding points in two neighboring frames. We determine the correspondence using the stabilized optical flow: $\mathbf{p}_2 = \mathbf{p}_1 + \hat{F}_{t}^j(\mathbf{p}_1)$. $\mathbf{\bar{F}}$ is the fundamental matrix between the two frames estimated by Least Median of Squares (LMedS) regression~\cite{lmeds}. 

\begin{figure*}[t]
  \centering

   \includegraphics[width=1\linewidth]{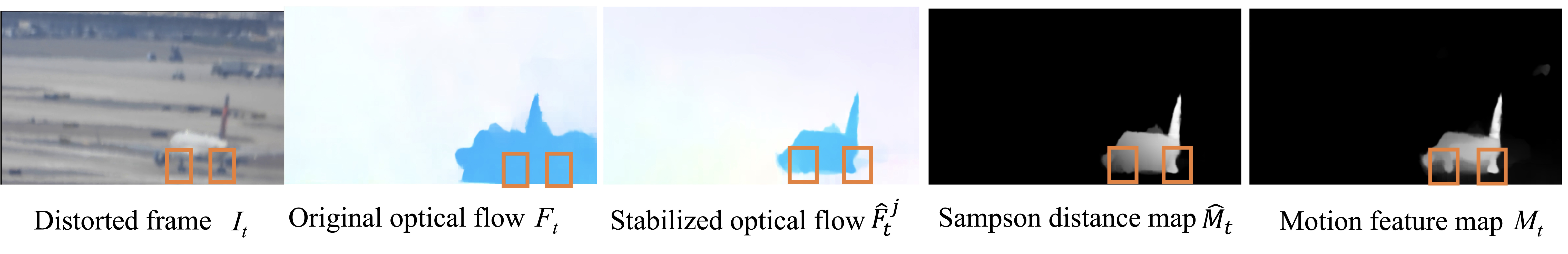}

   \caption{Step-by-step intermediate results for motion feature map estimation. }
   \label{fig:wheel}
   \vspace{-10pt}
\end{figure*}

We further average all available Sampson distance maps $\{\hat{M}_t^j | j = 1,2,..., A\}$ for a frame $I_t$ to obtain the per-frame motion feature map: $  {M}_{t} = \frac{1}{A} \sum_{j=1}^{A}  \hat{M}_{t}^j.$ This map indicates how likely a pixel is to belong to a moving object. Fig.~\ref{fig:wheel} compares the intermediate results when generating the motion feature map. Note that the original optical flow map is corrupted by turbulence and cannot resolve the airplane. After our stabilization and consistency check, the final motion feature map preserves the airplane's shape including the highlighted wheels.

\subsection{Region Growing-based Segmentation} \label{sec_region_grow}
\begin{figure*}[tb]
  \centering
    \includegraphics[width=1\linewidth]{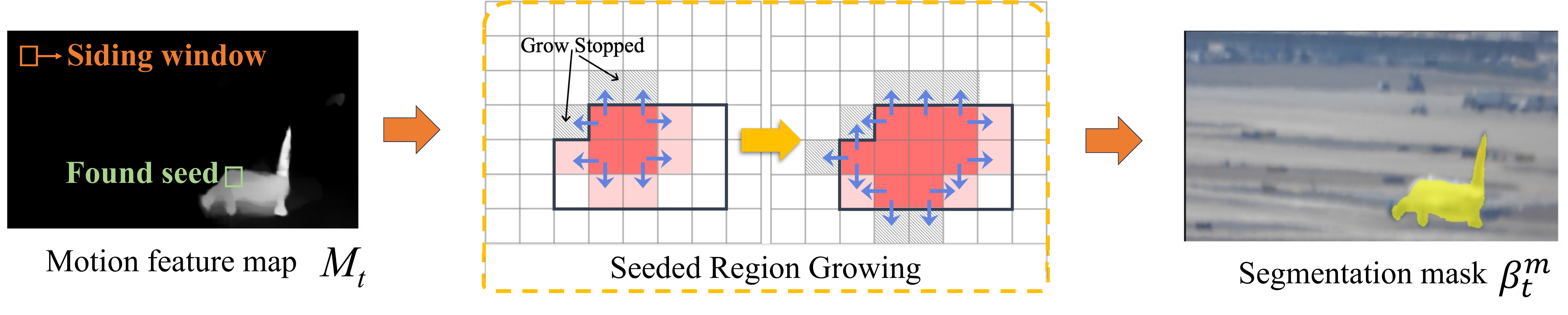}
    \caption{Pipeline of region growing-based segmentation. We select seeds on the motion feature map using a sliding window. We then grow the seeds to full segmentation masks.}
  \label{fig:regiongrow}
  \vspace{-10pt}
\end{figure*}

Next, based on the motion feature maps, we adopt a ``region growing'' scheme to generate segmentation masks for moving objects (see Fig.~\ref{fig:regiongrow}). Note that while motion feature maps effectively characterize object motions, they tend to be non-binary and exhibit fuzziness at the object boundaries.

\vspace{2pt}
\noindent\textbf{Initial Seed Selection.}
We first select a small set of seedling pixels that have high confidence of being on a moving object. This selection is based on the motion feature map which encodes how likely a pixel is to be in motion. 
Note that since we apply the sliding window on the motion map ${M}_{t}$, the size and appearance of the object would not affect the seed selection. 
Specifically, our sliding windows $\{W_k\}_{k=1}^K$ are of size $D\times D$, where $D$ is adaptively chosen based on the input resolution, to scan through $M_{t}$ to determine the initial seed of each moving object $k$. A seed is detected when pixels within the search window are of similar large values, indicating large Sampson distances in this area.
Specifically, for a search window $W_k$, we consider it has found a seed when it satisfies two criteria: (1) its average value $\bar{M_{t}}(W_k)$ is greater than a threshold $\delta_1$; and (2) its variance $\sigma^2(M_{t}(W_k))$ is less than a small threshold $\delta_2$. Each selected seedling region is assigned an integer mask ID to uniquely identify multiple moving objects.

\vspace{2pt}
\noindent\textbf{Seeded Region Growing.}
We then grow full segmentation masks from the initialized seedling pixels. We gradually expand each seeded region outwards to the nearest neighbors of boundary pixels using this criteria for pixel inclusion:
\begin{equation}
\label{eq:groweq}
    |M_t(\textbf{p}_{new})-M_t(\textbf{p}_{seed})| < \delta_{seed},
\end{equation}
where $M_t$ is the motion feature map; $\textbf{p}_{new}$ is the pixel under consideration; $\textbf{p}_{seed}$ is the seed pixel that we grow from; and $\delta_{seed} = 0.2\times M_t(\textbf{p}_{seed})$ is the threshold for stopping the growth. 
Note that this threshold value depends on turbulence strength and needs to be adjusted for extreme cases. 
For stronger turbulence, we prefer larger $\delta_{seed}$ and increase the multiplier from $0.2$ to $0.3$. For weak turbulence, we decrease the multiplier to $0.1$. The object's border tends to get blurred for videos with severe turbulence. Meanwhile, a larger threshold makes the region grow harder, resulting in more reliable results. 

When growing from multiple seeds, we skip the pixels already examined so that different object masks are non-overlapping. We thus obtain a set of per-frame segmentation masks $\{{\beta}^m_t\}_{t=1}^T$.

\vspace{2pt}

\noindent\textbf{Mask ID Unification.} When a scene has multiple moving objects (say $K$), our region-growing algorithm will generate $K$ segmentation masks with IDs ranging from 1 to $K$ for each frame. Since the region-growing module is applied to each frame independently, the mask ID among different frames may be inconsistent with respect to objects. There is a need to unify the mask IDs across frames, so that the same mask ID always maps to the same object. We propose a K-means-based filtering technique to do so. 

We first represent each mask region by its centroid: $\textbf{c}^m_t=\text{mean}(\textbf{p}^m_t)$ (where $\textbf{p}^m_t$ is the coordinates of all foreground pixels in the mask). We take the mask centroids for all frames and all objects, and optimize $K$ K-means cluster centroids $\mu^m$:
\begin{equation}
\underset{\mu^m}{\text{argmin}} \sum_{m=1}^{K} \sum_{t=1}^{T} ||\textbf{c}^m_t - \mu^m||^2,
\end{equation}
where K is the total number of objects, and T is the total number of frames.

We then re-ID the masks for each frame by comparing their centroids to the K-means cluster centroids. The K-means cluster IDs act as a global reference for all frames. Each mask is assigned the ID $m$ of a K-means cluster that it is closest to: 
$m^{*} = \underset{m}{\text{argmin}} ||\textbf{c}^m_t - \mu^m||.$
After this re-assignment, the mask IDs for all frames are consistent with each ID uniquely identifying a moving object in the scene. This allows our method to handle multiple moving objects in a scene. 

\subsection{Spatio-Temporal Refinement} \label{sec_refine}
Finally, we further refine the masks to improve their spatio-temporal consistency. We develop a Refine-Net $\Phi_{\theta}$ with a U-Net~\cite{10.1007/978-3-319-24574-4_28} backbone to refine the masks generated by region growing. 

\noindent\textbf{Parameter Initialization.}
We concatenate the video frame $I_t$ with its motion feature map $M_t$. The concatenated tensor is fed as input to the Refine-Net $\Phi_{\theta}$: $\alpha^m_t=\Phi_{\theta}(I_t, M_t)$ (here $\alpha^m_t$ is the output of $\Phi_{\theta}$, which is a refined mask). We use the following loss function for initializing the parameters of $\Phi_{\theta}$: 
\begin{equation}
  \mathcal{L}_{ini} = \gamma_1 \mathcal{L}_{1} + \gamma_2  \sum_{g}\mathcal{L}_{2}^g + \gamma_{3}  \sum_{g}\mathcal{L}_{3}^g,
  \label{eq: averaged optical flow}
\end{equation}
where $\gamma_1$, $\gamma_2$, and $\gamma_{3}$ are balancing weights for each loss term. We run $20-30$ epochs for initialization. Below, we describe our loss terms in detail. 

$\mathcal{L}_{1}$ is a pixel-wise cross-entropy loss that enforces consistency between the refined output mask $\alpha^m_t$ and coarse input mask $\beta^m_t$. It is calculated as:
\begin{equation}
\small
\begin{split}
  \mathcal{L}_{1} =  \frac{1}{\Omega}  \sum_{\textbf{p} \in \Omega} (-\alpha^m_t(\textbf{p})\log\alpha^m_t(\textbf{p}) +  \beta^m_t(\textbf{p})\log\beta^m_t(\textbf{p})),
  \label{eq: averaged optical flow}
\end{split}
\end{equation}
where $\textbf{p}$ is the pixel coordinates, and $\Omega$ represents the spatial domain.
%

$\mathcal{L}_{2}^g$ is a bidirectional consistency loss that enforces flow consistency between $\alpha^m_{t+g}$ and the optical flow-warped input mask: $\hat{\beta}^m_t={F}_{t \rightarrow t+g}(\beta_{t}^m)$.
We also use the cross-entropy for comparison, and $\mathcal{L}_{2}^g$ is written as: 
\begin{equation}
\small
\begin{split}
  \mathcal{L}_{2}^g=  \frac{1}{\Omega} \sum_{\textbf{p} \in \Omega} (-\alpha^m_{t+g}(\textbf{p})\log\alpha^m_{t+g}(\textbf{p}) + 
  \hat{\beta}^m_{t}(\textbf{p}) \log \hat{\beta}_{t}^m(\textbf{p})).
  \label{eq:l2}
\end{split}
\end{equation}
%

$\mathcal{L}_{3}^g$ is another bidirectional consistency loss that enforces flow consistency between $\alpha^m_{t+g}$ and the optical flow-warped version of itself: $\hat{\alpha}^m_t={F}_{t \rightarrow t+g}(\alpha_{t}^m)$. $\mathcal{L}_{3}^g$ is written as:
\begin{equation}
\small
\begin{split}
  \mathcal{L}_{3}^g=  \frac{1}{\Omega} \sum_{\textbf{p} \in \Omega} (-\alpha^m_{t+g}(\textbf{p})\log\alpha^m_{t+g}(\textbf{p}) + 
  \hat{\alpha}^m_{t}(\textbf{p}) \log \hat{\alpha}_{t}^m(\textbf{p})).
  \label{eq:l3}
\end{split}
\end{equation}

After initialization, the output mask $\alpha^m_t$ is aligned with the input mask $\beta^m_t$ and has improved temporal consistency with the two bidirectional losses.

\vspace{2pt}
\noindent\textbf{Iterative Refinement using Grouping Function.} 
To further improve the mask quality and consistency, we adopt an iterative refinement constrained by a grouping function. The refinement runs for 10 epochs. We update the input reference mask $\beta^m_t$ using a K-means-based grouping function for every 3 epochs.
Specifically, we concatenate each pixel's mask values $\{\beta^m_t(\textbf{p})\}_{t=1}^T$, with its coordinates $\textbf{p}=(x,y)$, to form a new tensor $\{T^m_t(\textbf{p})\}_{t=1}^T$ that combines both the motion and spatial information. The combined tensor $T^m_t(\textbf{p})$ of all pixels is used to optimize two K-means cluster centroids $\theta^1, \theta^2$: one for foreground cluster ($\theta^1$) and the other for background ($\theta^2$). The centroids are optimized using the following function:
\begin{equation}
\underset{\theta^1,\theta^2}{\text{argmin}} \sum_{i=1}^{2} \sum_{\textbf{p} \in \Omega}^{\Omega} ||T^i_t - \theta_t^i||^2,
\end{equation}
where $\Omega$ is the domain of all pixels. We then re-assign the values of $\beta^m_t$: if $T^m_t(\mathbf{p})$ is closer to $\theta^1$, we assign the mask value as 1; otherwise, we consider the pixel as background and assign the mask value as 0. 



The loss function we use for network optimization is the same as the initialization step, but in this refinement step, $\beta^m_t$ is updated using the K-means-based grouping every 3 epochs. Without this grouping-based refinement, the output mask tends to have gaps or other spatial inconsistent artifacts. 



\section{Experiments}

\begin{table}[tb]\scriptsize
  \caption{Comparison of existing datasets on turbulent images or videos. 
  }
  \label{tab:dataset}
  \centering
  \begin{tabular}{@{}c|c|c|c|c|c@{}}
    \toprule
    Dataset & Source & Format & Purpose & Num. of video & Availability \\
    \midrule
    \makecell[l]{Turb Pascal VOC~\cite{cui2019accurate}} & \makecell[c]{Synthetic} & Image & Segmentation & \makecell[c]{\XSolidBrush} & \makecell[c]{\XSolidBrush}\\
    
    \makecell[l]{Turb ADE20K~\cite{cui2019accurate}} & \makecell[c]{Synthetic} & Image & Segmentation & \makecell[c]{\XSolidBrush} & \makecell[c]{\XSolidBrush}\\
    
    \makecell[l]{TSRW-GAN~\cite{jin2021neutralizing}} & \makecell[c]{Real} & Video & Restoration & 27*  & \makecell[c]{\CheckmarkBold}\\
    
    \makecell[l]{OTIS~\cite{gilles2017open}} & \makecell[c]{Real} & Video & Restoration & 5   & \makecell[c]{\CheckmarkBold}\\
    
    \makecell[l]{BVI-CLEAR~\cite{CLEAR2013}} & \makecell[c]{Real} & Video & Restoration & 3  & \makecell[c]{\CheckmarkBold}\\

    \makecell[l]{DOST (ours)} & \makecell[c]{Real} & Video & Segmentation & \makecell[c]{38} & \makecell[c]{\CheckmarkBold}\\

  \bottomrule
  \end{tabular}
  
  \scriptsize{* The actual number of videos in TSRW-GAN is greater than 27, but many of the videos have large overlaps. The number of unique (or non-overlapped) videos is 27.}
\end{table}

\subsection{DOST Dataset}
We capture a long-range turbulent video dataset, which we call \textit{Dynamic Object Segmentation in Turbulence (DOST)}, to evaluate our method. DOST consists of 38 videos, all collected outdoors in hot weather using long focal length settings. All videos contain instances of moving objects, such as vehicles, aircraft, and pedestrians. We manually annotate the video to provide per-frame ground truth masks for segmenting moving objects. 
Specifically, we use a Nikon Coolpix P1000 to capture the videos. The camera has an adjustable focal length of up to $539$mm ($125\times$ optical zoom), which is equivalent to $3000$mm focal length in 35mm sensor format. We record videos with a resolution of $1920 \times 1080$. In total, our dataset has 38 videos with 1719 frames. We annotate moving objects in each video frame using the Computer Vision Annotation Tool (CVAT)\cite{cvat}. 
Our dataset is the first of its kind to provide a ground truth moving object segmentation mask in the context of long-range turbulent video. DOST is designed for motion segmentation, but can be used for other tasks (e.g., turbulent video restoration).  

Table \ref{tab:dataset} compares DOST with existing datasets designed for turbulent image or video processing. Since real images/videos with air turbulence are very difficult to acquire, real datasets are scarce, and existing ones are usually small in size. 
Cui and Zhang~\cite{cui2019accurate} synthesize large turbulent image datasets using standard image datasets (Pascal VOC 2012\cite{cui2019accurate} and ADE20K\cite{cui2019accurate}) and a turbulence simulator. However, there is a domain gap between simulated data and real data. Further, their datasets only contain single images and cannot be used for studying motion. 
Other real turbulent video datasets~\cite{CLEAR2013,gilles2017open,jin2021neutralizing} are all designed for the restoration task. Although some~\cite{gilles2017open,jin2021neutralizing} have bounding box annotations, none provide object-tight segmentation masks.

\subsection{Implementation Details}
We implemented our network using PyTorch on a supercomputing node equipped with an NVIDIA GTX A100 GPU. The input frames are resized to a lower resolution of $240 \times 432$ for faster optical flow calculation and network training. We use RAFT \cite{teed2020raft} for optical flow estimation with a maximum frame interval of 4. The stabilized optical flow and Sampson distance maps are subsequently calculated based on the RAFT output. The region-growing algorithm's stopping threshold $\delta_{seed}$ (see Eq.~\ref{eq:groweq}) is dependent on the turbulence strength, and we need to adjust this parameter for varied strength turbulent videos (see more details in the supplementary material). For the bidirectional consistency losses, we compare with four neighboring time frames, $g \in \{-2, -1, 1, 2\}$ in Eqs.~\ref{eq:l2}-\ref{eq:l3}.  
\vspace{2pt}

\noindent\textbf{Evaluation Metrics.}
Given a ground-truth mask, we evaluate the accuracy of the estimated segmentation mask using two standard metrics~\cite{10350598}: (1) Jaccard's Index ($\mathcal{J}$) that calculates the intersection over the union of two sets (also known as intersection-over-union or IoU measure); and (2) F1-Score ($\mathcal{F}$) that calculates the harmonic mean of precision and recall (also known as dice coefficient). We also calculate the average of $\mathcal{J}$ and $\mathcal{F}$, and denote this overall metric as $\mathcal{G}$.

\vspace{2pt}

\subsection{Comparison with State-of-the-art Methods}






\begin{table}[tb]\scriptsize
  \caption{Quantitative comparisons with state-of-the-art unsupervised methods on DOST w.r.t. various turbulence strengths. 
  }
  \label{table2}
  \centering
  \begin{tabular}{@{}l|ccc|ccc|c@{}}
    \toprule
      &  & $\mathcal{J}$ &  &  & \makecell[c]{$\mathcal{F}$} &   & \makecell[c]{$\mathcal{G}$} \\
    
    Model  & \makecell[c]{Normal turb.}  & \makecell[c]{Severe turb.} & \makecell[c]{Overall} & \makecell[c]{Normal turb.} & \makecell[c]{Severe turb.} & \makecell[c]{Overall} & \makecell[c]{Overall} \\
    \midrule
    TMO~\cite{cho2023treating}  & 0.643  & 0.235  & 0.439  & 0.757  & 0.315  & 0.536 & 0.487\\

DSprites~\cite{ye2022deformable}  & 0.427  & 0.101  & 0.264  & 0.772  & 0.203  & 0.374 & 0.319\\

DS-net~\cite{LIU2022103700}  & 0.361  & 0.191  & 0.276  & 0.422  & 0.232  & 0.327 & 0.302\\

\midrule

Ours  & \textbf{0.851 $\uparrow$}  & \textbf{0.557 $\uparrow$}  & \textbf{0.703 $\uparrow$}  & \textbf{0.812 $\uparrow$}  & \textbf{0.634 $\uparrow$}  & \textbf{0.723 $\uparrow$} & \textbf{0.713 $\uparrow$}\\
  \bottomrule
  \end{tabular}
\end{table}

We compare our method against recent state-of-the-art \textit{unsupervised} methods for motion segmentation, including TMO~\cite{cho2023treating}, Deformable Sprites~\cite{ye2022deformable}, and DS-Net~\cite{LIU2022103700}. TMO~\cite{cho2023treating} achieves high accuracy in object segmentation regardless of motion. It therefore has certain advantages in handling turbulent videos, although it is not specifically designed for this purpose. Deformable Sprites~\cite{ye2022deformable} integrates appearance features with optical flow, and further enforces consistency with optical flow-guided grouping loss and warping loss. DS-Net~\cite{LIU2022103700} uses multi-scale spatial and temporal features for segmentation and is able to achieve good performance when the input is noisy. For all three methods, we use the code implementations provided by authors and train their networks using settings described in papers. For methods that need optical flow, we use RAFT to estimate optical flow between consecutive frames. 


\vspace{2pt}
\noindent\textbf{Quantitative Comparisons.} We show quantitative comparison results in Table \ref{table2}. All methods are evaluated on our DOST dataset. We organize videos into two sets, ``normal turb.'' and ``severe turb.'', according to their exhibited turbulence strength. Our method significantly outperforms these state-of-the-art on motion segmentation accuracy under various turbulence strengths. In normal cases, some can still achieve decent performance, whereas our method scores much higher in all metrics. Compared to TMO, whose overall score is the highest among the three state-of-the-art, our accuracy is increasing by 60.1$\%$ in $\mathcal{J}$ and 34.9$\%$ in $\mathcal{F}$.
In severe cases, the performance of all state-of-the-art significantly downgrades, with all $\mathcal{J}$ values lower than 0.25 and $\mathcal{F}$ lower than 0.35. In contrast, our method is relatively robust to strong turbulence. 

\begin{figure}[tb]
  \centering
  \begin{subfigure}{0.53\linewidth}
    \includegraphics[width=1\linewidth]{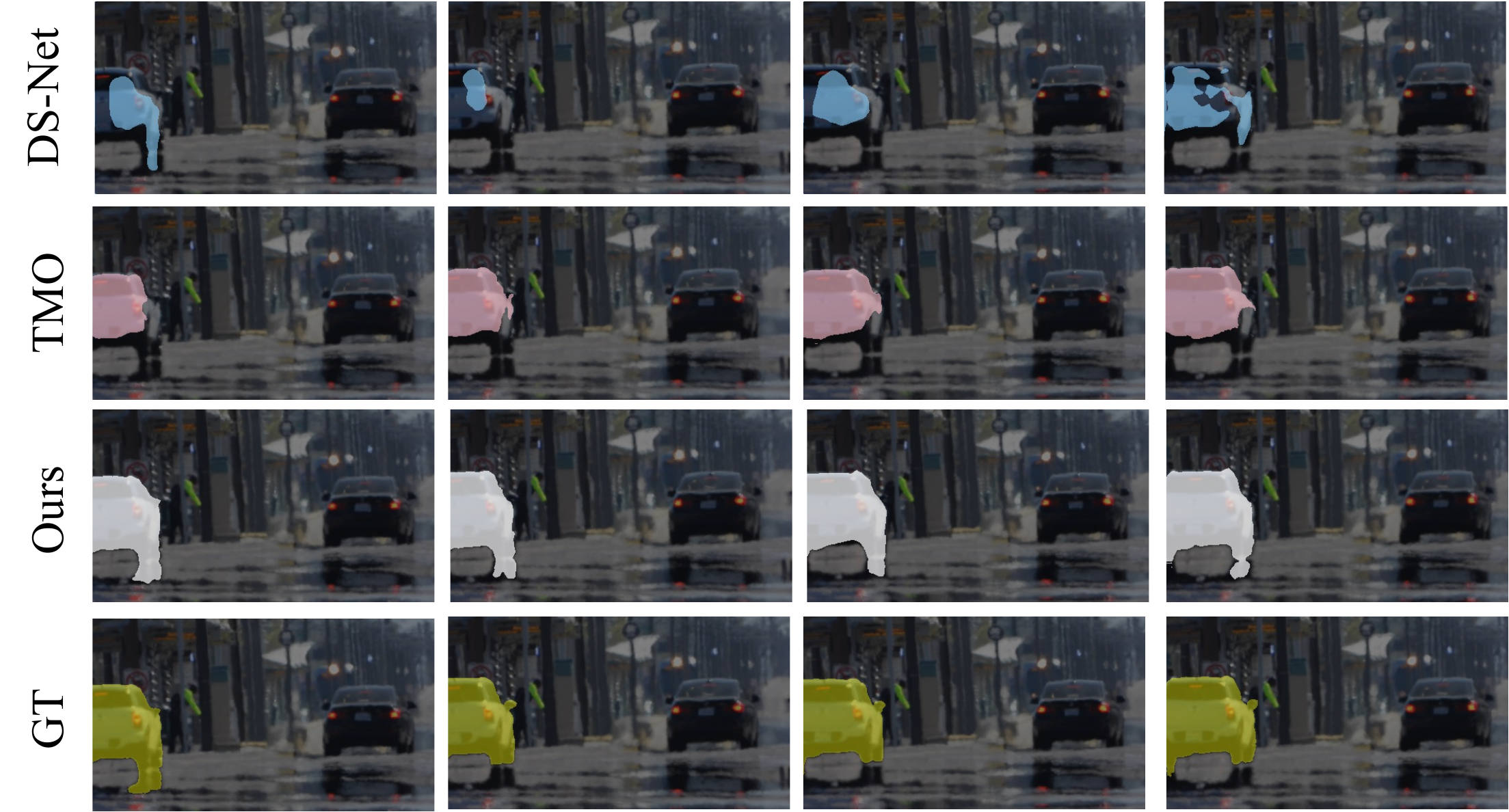}
    \caption{Comparisons w.r.t. camera shake.}
    \label{fig:a}
  \end{subfigure}
  \hfill
  \begin{subfigure}{0.45\linewidth}
    \includegraphics[width=0.9\linewidth]{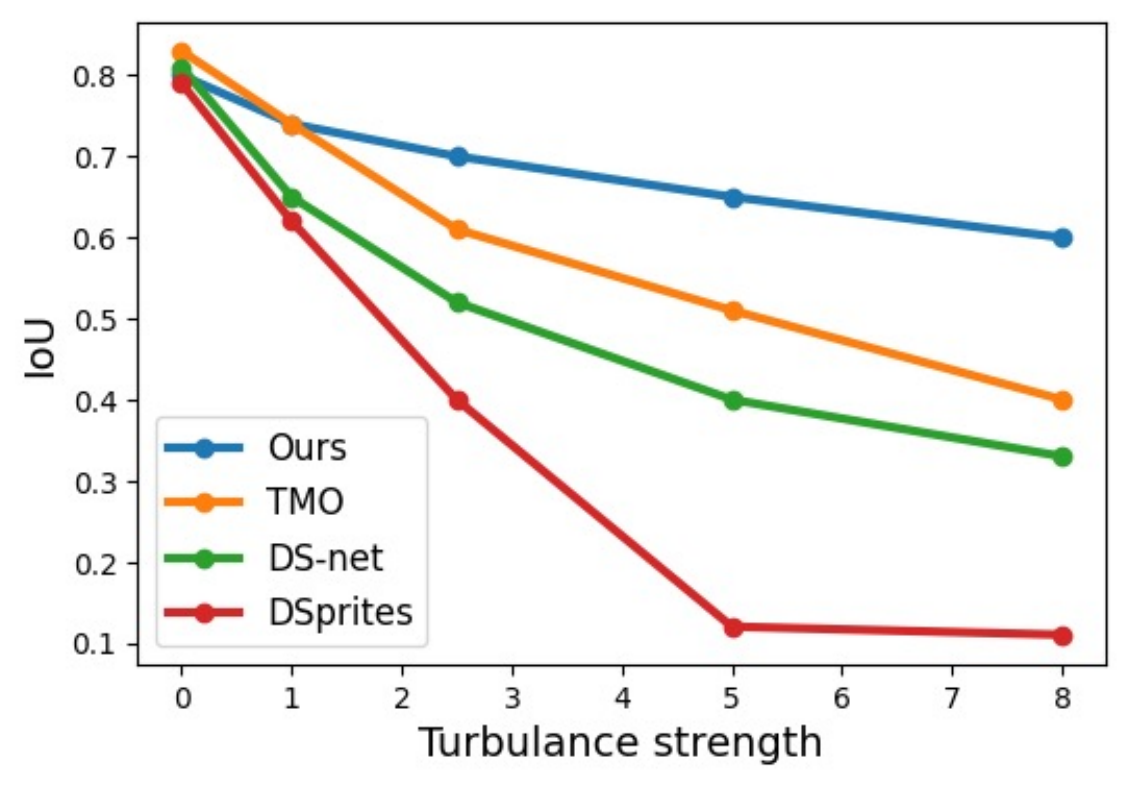}
    \caption{Comparisons w.r.t. turbulence strength.}
    \label{fig:b}
  \end{subfigure}
  
  \caption{Example results of handling videos (a) with camera motion and (b) impacted by different turbulence strength. (a) Our results achieves robust performance on different time frames (2nd to 4th columns) in videos suffering from significant camera shake. (b) We can also tell that our method achieves the best robustness across different turbulence strengths, even when it is very strong.}
  \label{fig:short}
\label{fig:stable}
\end{figure}

We also experiment on a larger synthetic dataset to evaluate our robustness with respect to turbulence strength. We use a physics-based turbulence simulator \cite{9711075} and the DAVIS 2016 dataset \cite{Perazzi2016} to synthesize videos with various strengths of turbulence. We test the three state-of-the-art methods on the synthetic data as well. $\mathcal{J}$ score (or IoU) plots with respect to turbulence strength are shown in Fig.~\ref{fig:b}. The results demonstrate that our method is robust to various turbulence strengths. Note that when there is no turbulence in the scene (strength = 0), TMO has a slightly higher $\mathcal{J}$ score, as it incorporates more visual cues for segmentation, whereas our method focuses on turbulence artifacts. 

\begin{figure*}[t]
\centering
\includegraphics[width=1\textwidth]{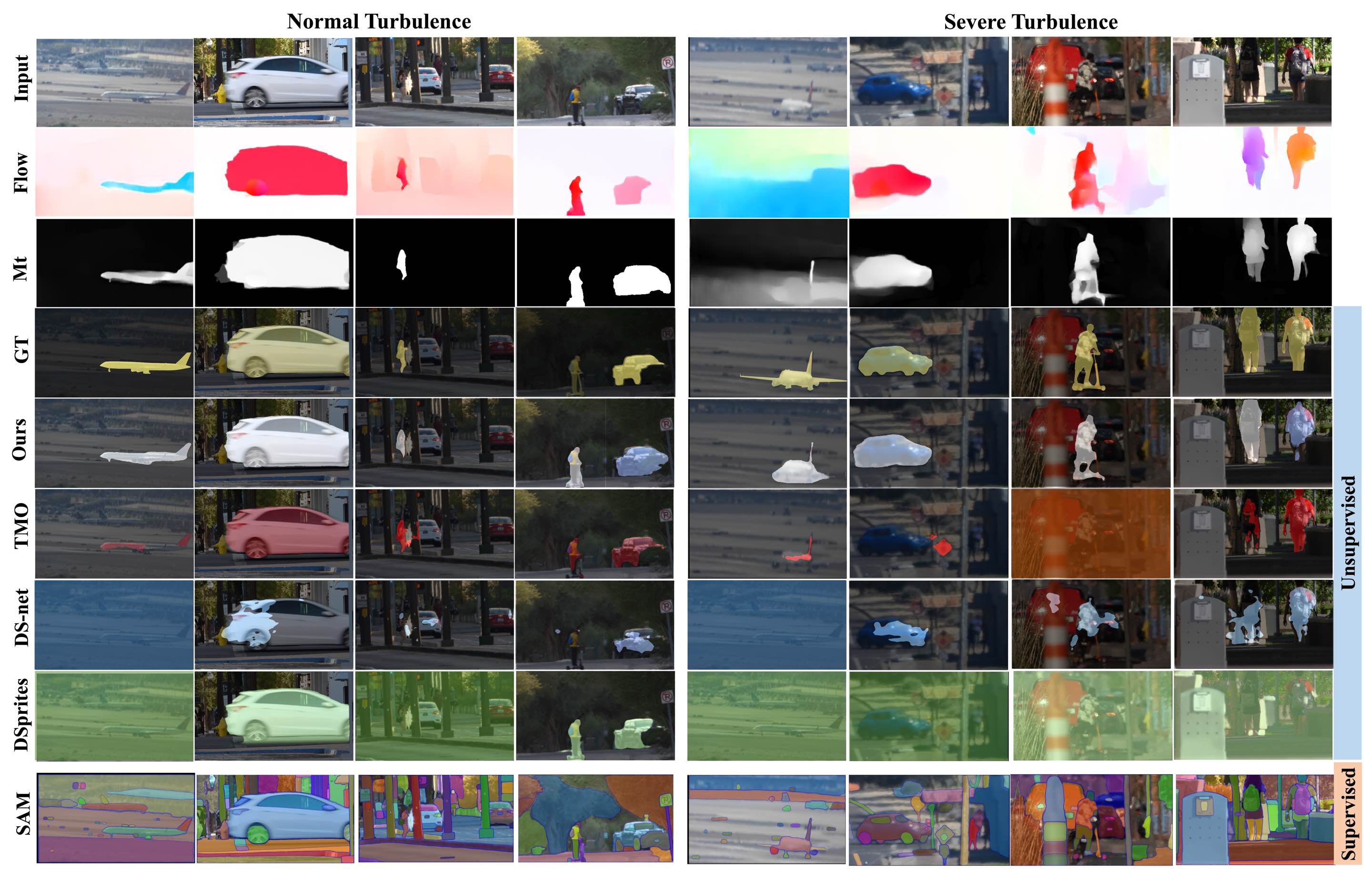} 
\caption{Qualitative comparisons with state-of-the-art methods on DOST w.r.t. various turbulence strengths. Here we also show the raw optical flows (``Flow''), our motion feature maps (``$M_t$''), and our ground truth masks (``GT'').}
\label{Qualitative comparison}
\vspace{-10pt}
\end{figure*}

\vspace{2pt}
\noindent\textbf{Qualitative Comparisons.}
We show visual comparisons with state-of-the-art in Fig.~\ref{Qualitative comparison}. We can see that the optical estimations are largely affected by turbulence, especially in severe cases. Our method is able to generate segmentation masks that are tight to the object and works well for multiple moving objects. State-of-the-art methods face challenges when the turbulence strength is strong, or the moving object is too small. Their segmentation masks are incomplete in many cases. In the airplane scenes, DS-Net and DSprites fail to detect moving objects.
Notably, our method achieves the highest robustness to turbulent distortions, and camera shakes, as shown in Fig.~\ref{fig:stable}. 
We analyzed the effects of various camera motions, including complex multi-directional shake, on segmentation results in video sequences, as illustrated in Fig.~\ref{fig:a}.
Quantitatively, our method achieved an average IoU score of 0.712 on videos featured by camera shaking in our dataset, compared with TMO/0.305, DS-Net/0.267, and DSprites/0.235, respectively. 

We notice that supervised segmentation methods can hardly generalize to DOST, since these methods were trained solely using turbulence-free data. For instance, the foundation model for segmentation, i.e., SAM ~\cite{kirillov2023segment}, which has been trained on 11 million images, and over 1B masks, still fails at segmenting whole objects under strong turbulence, as shown in Fig.~\ref{Qualitative comparison} (last row). Additionally, SAM does not interpret motion , and SAM requires a user to click or prompt the algorithm, whereas we only segment moving objects without any need for user input. More comparison results can be found in supplemental material. 




\begin{figure}[tb]
  \centering
  \begin{subfigure}{0.5\linewidth}
    \includegraphics[width=0.9\linewidth]{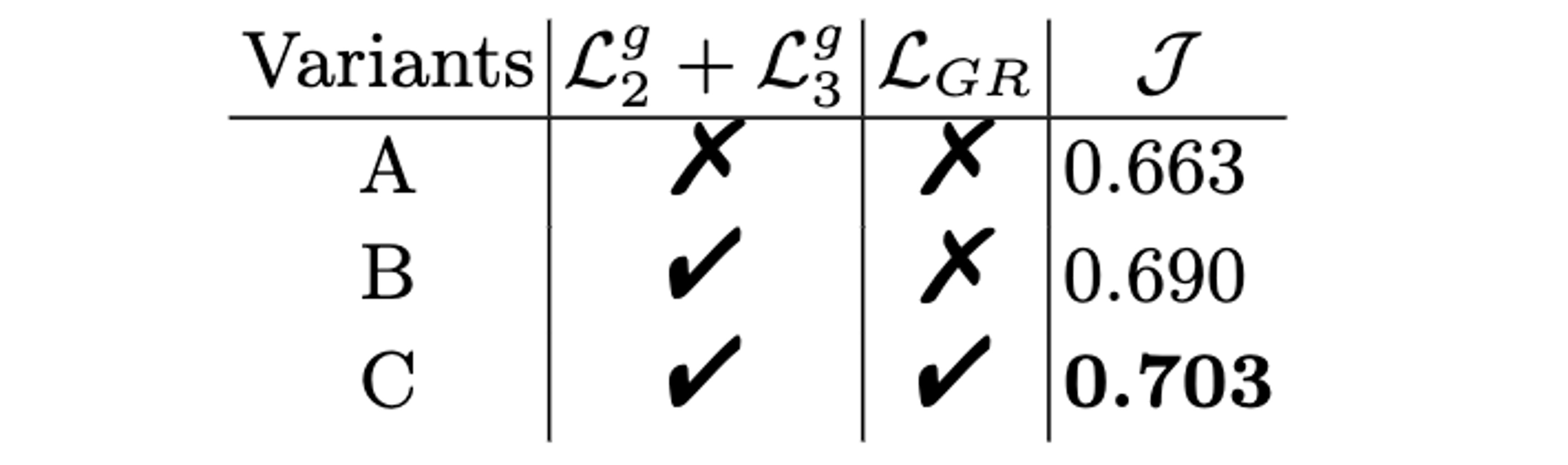}
    \caption{Ablation studies on our key components.}
    \label{fig:ablation1}
  \end{subfigure}
  \hfill
  \begin{subfigure}{0.43\linewidth}
    \includegraphics[width=0.9\linewidth]{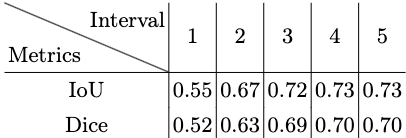}
    \caption{Ablation study on optical flow interval. }
    \label{fig:ablation2}
  \end{subfigure}
  \caption{Ablation studies. (a) Variants of our network. $A$: Simply using region-growing; $B$: Full pipeline without grouping loss; $C$: Our full pipeline. (b) The interval $i=1...5$ is the maximum temporal gap we used to stablize optical flow in Eqn.\ref{averaged optical flow}.}
  \label{fig:ablation}
\end{figure}

\subsection{Ablation Studies}
We perform ablation studies to evaluate individual components of our method. All experiments are performed on DOST. We test on three variants of our method: A only employs the region-growing algorithm (with Refine-Net excluded); B uses both region-growing and Refine-Net, but excludes the grouping loss for refinement; and C is implemented as our full approach. Their $\mathcal{J}$ score comparison results are shown in Fig. \ref{fig:ablation1}. Each component is clearly effective, and our full model achieves the best performance. We also evaluate the influence of the optical flow interval (i.e., the number of temporal frames used for optical flow calculation). Accuracy scores with respect to the interval length are shown in Fig.~\ref{fig:ablation2}. We can see that the score achieves the plateau when the interval is greater than 4. Therefore, we set the interval to 4. 

We also evaluated the effectiveness of our optical flow stabilization and geometric consistency check. Without the optical flow stabilization step, the IoU immediately drops to 0.354; without the geometric consistency check step, the IoU is 0.685, compared with IoU of 0.703 in our full pipeline.

\section{Conclusions}

In summary, we present an unsupervised approach for segmenting moving objects in videos affected by air turbulence. Our method uses a geometry-based consistency check to disambiguate motions and a region-growing scheme to generate tight segmentation masks. The masks are further refined with spatio-temporal consistency losses. Our method significantly outperforms the existing state-of-the-art in terms of accuracy and robustness when handling turbulent videos. We also contribute the first long-range turbulent video dataset designed for motion segmentation. 
Nevertheless, due to the unsupervised nature of our method, the current version can only achieve a latency of 0.95 FPS. Our future work will focus on optimizing the approach to reduce this latency by embedding results from foundations models such as SAM. Our method also has limited performance in separating overlapping moving objects in the videos. To address this, we plan to integrate additional visual cues, such as appearance and saliency.


\section*{Acknowledgements}
This research is based upon work supported in part by the Office of the Director of National Intelligence (ODNI), Intelligence Advanced Research Projects Activity (IARPA), via 2022-21102100003 and NSF IIS-2232298/2232299/2232300. The views and conclusions contained herein are those of the authors and should not be interpreted as necessarily representing the official policies, either expressed or implied, of ODNI, IARPA, NSF or the U.S. Government. The U.S. Government is authorized to reproduce and distribute reprints for governmental purposes, notwithstanding any copyright annotation therein. We also wish to thank ASU Research Computing for providing GPU resources~\cite{SOL} to support this research.

%
%
\bibliographystyle{splncs04}
\bibliography{main}
\end{document}